\documentclass{article} %
\usepackage{iclr2027_conference,times}

\usepackage{amsmath,amsfonts,bm}

\def\eqref#1{equation~\ref{#1}}

\def\1{\bm{1}}

\DeclareMathAlphabet{\mathsfit}{\encodingdefault}{\sfdefault}{m}{sl}
\SetMathAlphabet{\mathsfit}{bold}{\encodingdefault}{\sfdefault}{bx}{n}

\usepackage{amsmath,amssymb}
\usepackage{booktabs}
\usepackage{graphicx}
\usepackage{xcolor}
\usepackage{tikz}
\usepackage{pgfplots}
\pgfplotsset{compat=1.18}
\usepackage{wrapfig}
\usepackage{enumitem}
\usetikzlibrary{positioning,arrows.meta,fit,backgrounds,calc}
\usepackage[colorlinks=true,allcolors=blue!55!black]{hyperref}
\usepackage{url}

\definecolor{cobs}{HTML}{2F6FB5}   %
\definecolor{cgoal}{HTML}{0E8A6E}  %
\definecolor{cval}{HTML}{C1651A}   %
\definecolor{cact}{HTML}{6B3FA0}   %
\definecolor{cgray}{HTML}{5B6470}
\definecolor{closs}{HTML}{9A3324}

\newcommand{\method}{GlanceWAM}

\iclrfinalcopy
\title{GlanceWAM: Sparse Test-Time Imagination for
World-Action Models}

\author{
Linhan Wang$^{1}$ \quad Zijian An$^{2}$ \quad Mingyuan Zhang$^{3}$ \quad Chen Dai$^{1}$ \quad Yi Xu$^{3}$ \quad Can Cui$^{4}$ \\
\bf Jiayan Wang$^{3}$ \quad Zichong Yang$^{4}$ \quad Yinlin Chen$^{1}$ \quad Lifeng Zhou$^{2}$ \quad Chang-Tien Lu$^{1}$ \\[3pt]
\normalfont $^{1}$Virginia Tech \quad $^{2}$Drexel University \quad $^{3}$Northeastern University \quad $^{4}$Purdue University
}

\begin{document}
\maketitle
\lhead{Preprint.}

\begin{abstract}
Video generative models provide rich physical priors for robot learning, yet
existing world-action models (WAMs) face a fundamental trade-off: synchronous
video generation at control rate is latency-prohibitive, while abandoning
test-time visual imagination sacrifices task success.
We show that visual imagination achieves both real-time inference and
superior success rates when generated asynchronously off the critical path and consumed directly
in latent space.
We introduce \method{}, which decouples imagination from control on a
single shared video DiT backbone: an asynchronous proposer glances ahead on a slow clock to imagine a single
lookahead frame seconds into the future in the background, while an action head decodes
action chunks at control rate ($48$\,ms) purely in latent space without blocking. Enabled by a
non-interfering attention mask that isolates video representations and
staleness-robust horizon training that accommodates asynchronous lookahead
aging, \method{} breaks the speed--success dilemma. Trained purely on demonstrations,
it attains $72.2\%$ on the 24-task RoboCasa kitchen benchmark (vs.\ $67.1\%$ for
synchronous Cosmos Policy) and $99.0\%$ on LIBERO while cutting per-chunk control latency $24\times$
relative to synchronous world-action models ($48$\,ms on one A100). In single-arm and bimanual real-robot manipulation, it
achieves higher average success than $\pi_{0.5}$ without any robot-data pretraining.
Code is available at \url{https://github.com/linhanwang/GlanceWAM}; real-robot videos are on the project page \url{https://linhanwang.github.io/glancewam/}.
\end{abstract}

\section{Introduction}
\label{sec:intro}

\begin{figure}[t]
\centering
\includegraphics[width=\textwidth]{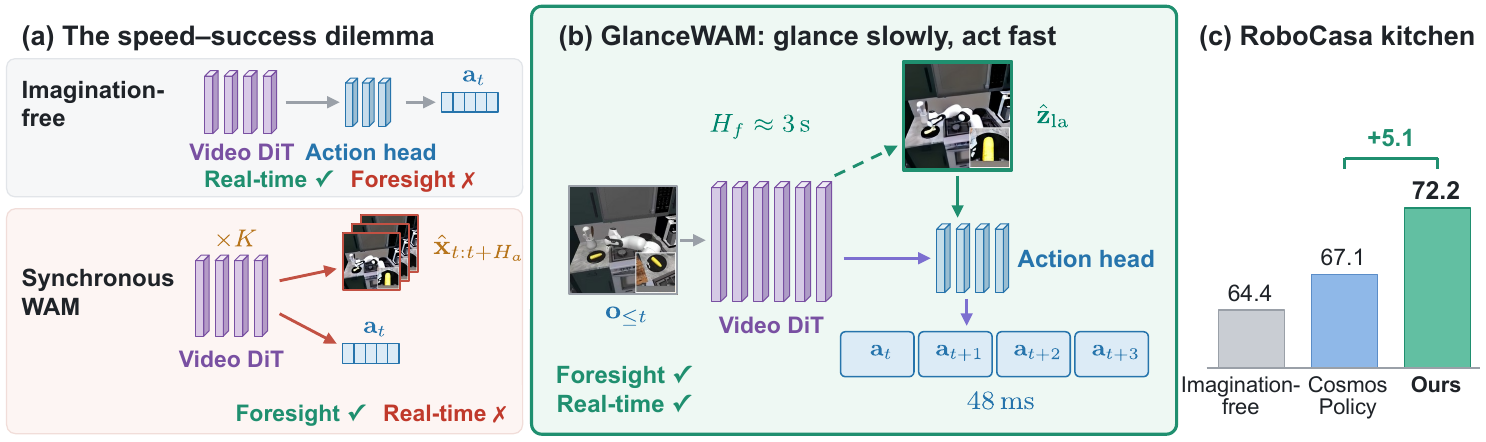}
\caption{\textbf{Synchronous imagination versus sparse lookahead foresight.}
\textbf{(a)}~Existing WAMs either skip test-time foresight or imagine video synchronously with every action chunk, paying $K$ heavy video denoising steps per chunk.
\textbf{(b)}~\method{} glances ahead asynchronously to a single latent lookahead frame $H_f\approx3$\,s away and reuses it across ${\approx}4$ action chunks, each decoded in $48$\,ms.
\textbf{(c)}~Success rate on RoboCasa kitchen (24 tasks, demos only).}
\label{fig:teaser}
\end{figure}

Video generative models capture rich physical priors over object dynamics, contact physics, and 3D scene evolution~\citep{cosmos2025,wan2025,skyreels2025}, offering a promising foundation for autonomous robots to anticipate the consequences of their actions. In robot manipulation, world-action models (WAMs) leverage these predictive backbones through two distinct pathways: \emph{(i)}~\emph{representation shaping}, where future-prediction losses enrich shared visual features, and \emph{(ii)}~\emph{visual foresight}, where the model explicitly synthesizes future visual states to guide downstream policy execution~\citep{cosmospolicy2026}. However, existing WAMs couple future prediction and action decoding synchronously at the control rate (Figure~\ref{fig:teaser}a)~\citep{cosmospolicy2026,lingbotva2026}. This tight coupling incurs two fundamental costs: prohibitive inference latency and myopic foresight. Over the short duration of an action chunk the scene barely changes, so predicting only that far ahead reduces world modeling to near-trivial reconstruction of the current observation.

To bypass this latency bottleneck, recent approaches advocate abandoning test-time visual imagination entirely, relegating video modeling purely to offline pretraining or feature regularization~\citep{fastwam2026,ahawam2026}. While this strategy eliminates generative overhead during control, it discards explicit visual foresight---depriving downstream policies of the future visual destinations and spatial guidance needed for long-horizon manipulation. Conversely, architectures that retain visual foresight continue to couple dense future synthesis directly to high-frequency action chunks~\citep{cosmospolicy2026,lingbotva2026}, remaining bounded by prohibitive diffusion latencies and near-static prediction horizons. Current world-action models are thus caught in a \emph{speed--success dilemma}: either synthesize video synchronously at the control rate and forfeit real-time reactivity, or discard test-time imagination and forfeit the performance benefits of visual foresight.

In this paper, we show that this dilemma is not an intrinsic property of visual foresight, but an artifact of coupling imagination synchronously to high-frequency control chunks. Downstream policies do not require dense, frame-by-frame future video of the immediate milliseconds; they primarily need a distant spatial destination (``where to go''). We introduce \method{}, a framework that decouples visual imagination from real-time control on a single shared video DiT backbone. \method{} makes visual foresight fast through asynchronous execution off the critical path: the model glances ahead on a slow clock to generate a single lookahead frame seconds into the future ($H_f \approx 3$\,s) in the background, completely isolating generative sampling latency from the high-frequency control loop. Crucially, the entire imagination-and-control loop operates purely in latent space without decoding to raw pixels, allowing the action head to decode short action chunks at the control rate ($48$\,ms) in a single forward pass. This architecture is enabled by two key mechanisms: \emph{(i)}~a non-interfering attention mask (prefix-LM) that isolates video representations by preventing lookahead tokens from contaminating observation encodings, and \emph{(ii)}~staleness-robust horizon training that supervises the policy across varying time offsets ($u \sim \mathcal{U}(0, H_f]$) to tolerate lookahead aging.

Trained purely on demonstrations, \method{} establishes a new state-of-the-art among world-action models across both manipulation success rate and inference speed, resolving the speed--success dilemma in practice. On the 24-task RoboCasa kitchen benchmark~\citep{robocasa2024}, \method{} achieves $72.2\%$ success, outperforming both synchronous Cosmos Policy ($67.1\%$) and imagination-free co-training ($64.4\%$), while reaching $99.0\%$ on LIBERO~\citep{libero2023}. Concurrently, its action decoding executes in $48$\,ms per chunk on a single NVIDIA A100 GPU ($24\times$ lower control-path latency than synchronous WAMs), operating comfortably within real-time control budgets. In single-arm and bimanual real-robot manipulation, \method{} achieves higher average success than $\pi_{0.5}$~\citep{pi05_2025} without any robot-data pretraining (\S\ref{sec:real}). Systematic diagnostics further confirm that lookahead conditioning is causally load-bearing and remains robust under asynchronous execution delays, demonstrating that world-action models do not need to choose between speed and imagination---sparse, asynchronous visual foresight delivers both.

\section{Related work}
\label{sec:related}

\textbf{World-action models.}
Generative video models capture expressive physical priors over scene dynamics, prompting diverse strategies to harness them for robot manipulation. Early ``predict-then-act'' paradigms synthesize dense multi-frame video rollouts and decode actions through inverse dynamics~\citep{unipi2023,hip2023,turningvideo2026}, incurring severe generative latency on the critical control path. Recent world-action models (WAMs) tighten this integration: Cosmos Policy~\citep{cosmospolicy2026} co-denoises robot actions, single future observation frames, and value estimates within a unified diffusion sequence; LingBot-VA~\citep{lingbotva2026}, VideoVLA~\citep{videovla2025}, and GigaWorld-Policy~\citep{gigaworld2026} autoregressively interleave video and action tokens across matched intervals; and UWM~\citep{uwm2025} jointly trains video and action diffusion. In these architectures, future prediction is tightly coupled to the action chunk horizon (e.g., $0.8$--$1.6$\,s), paying multi-step diffusion sampling overhead on every action chunk while predicting near-static short-horizon transitions. Conversely, purely auxiliary frameworks such as FLARE~\citep{flare2025} leverage future prediction strictly as a representation learning objective without test-time foresight. \method{} unifies both pathways: it decouples timescales to generate seconds-scale lookaheads ($H_f \approx 3$\,s) on an amortized slow clock, while retaining real-time $48$\,ms action decoding purely in latent space. Concurrent works such as DeVA~\citep{deva2026} and Flex-$\pi$~\citep{flexpi2026} explore alternative joint video--action denoising and compute-flexible architectures (detailed comparisons in Appendix~\ref{app:concurrent}).

\textbf{Test-time imagination and efficiency.}
A recent line of inquiry questions whether world-action models require test-time visual imagination at all. Fast-WAM~\citep{fastwam2026} and AHA-WAM~\citep{ahawam2026} argue that future prediction is unnecessary during inference and can be relegated entirely to offline representation shaping. Both achieve this through causal attention masking, isolating observation token encodings so that future prediction tokens can be losslessly removed at test time. AHA-WAM further introduces observation-guided context routing and phase-offset training to refresh stale planner representations under temporal latency. Complementary acceleration efforts explore progressive distillation~\citep{flashwam2026}, lightweight 1B architectures~\citep{efficientwam2026,lightwam2026}, persistent rolling memory~\citep{memorywam2026}, or predictive representation folding~\citep{enfold2026}. While removing or distilling test-time generation mitigates sampling latency, amortizing world dynamics strictly into static weights deprives downstream policies of explicit visual targets. In this work, we demonstrate within a controlled, unified architecture that test-time visual foresight provides critical task guidance ($+7.8\%$ on RoboCasa kitchen), and show that sparse asynchronous amortization removes the latency bottleneck without discarding imagination. Unlike prior WAMs that retain test-time imagination, \method{} takes generation off the control path.

\textbf{Visual lookaheads and foresight.}
Conditioning visuomotor policies on future visual targets has a rich foundation in hierarchical robot learning. Prior methods instantiate visual subgoals through image-editing models~\citep{susie2024}, progress-filtered subgoal candidates~\citep{ghilglue2025}, and dedicated high-level video planners refreshed every few seconds~\citep{pi07_2026}. While \method{} shares the principle of multi-second visual lookahead conditioning, it unifies the lookahead generator and policy backbone within a single video DiT rather than maintaining separate, disjoint models (e.g., BAGEL and $\pi_0$ in~\citet{pi07_2026}), ensuring that predictive world modeling directly shapes shared policy representations. On latency and representation grounds, an emerging line explores non-RGB and latent-space foresight: LaWAM~\citep{lawam2026} and RepWAM~\citep{repwam2026} propose predicting latent visual features or representation tokenizers rather than raw pixels; EgoWAM~\citep{egowam2026}, DreamWAM~\citep{dreamwam2026}, LiLa-WAM~\citep{lilawam2026}, and VLA-JEPA~\citep{vlajepa2026} explore geometry, 3D flow, and joint-embedding predictive representations. However, in methods like LaWAM, subgoals are generated conditioned on the policy's already-predicted actions, rendering the subgoal downstream of action selection. In contrast, \method{} synthesizes action-independent lookahead latents seconds in advance to provide explicit spatial destinations (``where to go'') that guide subsequent action chunks directly in latent space.

\section{Method}
\label{sec:method}

In this section, we present \method{}, a world-action model that decouples test-time visual foresight from high-frequency action execution on one shared video DiT backbone, read by a lightweight flow-matching action head~\citep{flowmatching2023,gr00t2025}.
We formalize the dual-timescale problem (\S\ref{sec:preliminaries}), then detail the architecture (\S\ref{sec:architecture}), staleness-robust co-training (\S\ref{sec:training}), and asynchronous latent-space inference (\S\ref{sec:inference}).

\begin{figure}[t]
\centering
\includegraphics[width=0.98\textwidth]{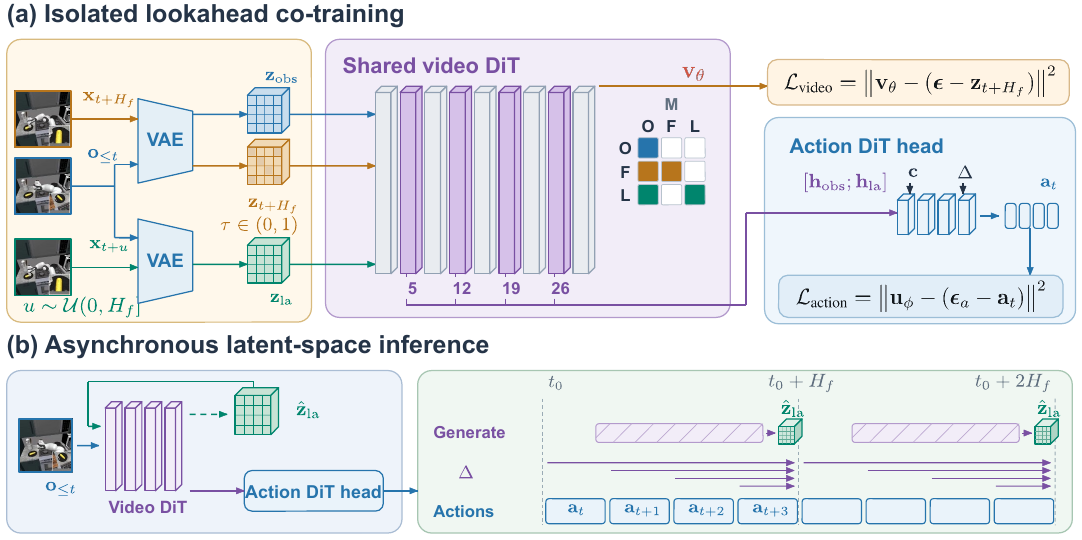}
\caption{\textbf{\method{} architecture.} \textbf{(a)}~In co-training, the video DiT predicts the future target $\mathbf{x}_{t+H_f}$, while the action head learns inverse dynamics toward a lookahead frame $\mathbf{x}_{t+u}$ at a random offset $u \sim \mathcal{U}(0, H_f]$, conditioned on $\Delta = u$, so it sees every lookahead age it will meet at deployment. Two separate VAE passes and the prefix-LM mask $\mathbf{M}$ keep the lookahead out of video prediction. \textbf{(b)}~At inference, the lookahead latent $\hat{\mathbf{z}}_{\text{la}}$ is generated asynchronously once per $H_f$ and never decoded; each $0.8$\,s action chunk reads the held $\hat{\mathbf{z}}_{\text{la}}$ at the decaying offset $\Delta$ in $48$\,ms, within the offset range seen in training.}
\label{fig:method}
\end{figure}

\subsection{Problem Formulation and Dual-Timescale Setup}
\label{sec:preliminaries}

\textbf{Setting and video diffusion foundation.}
We consider language-conditioned visuomotor manipulation from demonstration trajectories.
At each decision step $t$, the policy receives observation history $\mathbf{o}_{\le t}$ and instruction $\mathbf{c}$, predicting an action chunk $\mathbf{a}_t = [a_t, \dots, a_{t+H_a-1}] \in \mathbb{R}^{H_a \times D_a}$ spanning control horizon $H_a$ ($16$ steps at $20\,\text{Hz} = 0.8\,\text{s}$)~\citep{act2023,diffusionpolicy2023,cosmospolicy2026}.
Our framework builds upon a latent video diffusion transformer (SkyReels-V2-DF, 1.3B)~\citep{skyreels2025} that compresses video frames into latent representations $\mathbf{z}_i$ via a frozen causal video VAE encoder $\mathcal{E}$~\citep{wan2025} (reproducibility details in Appendix~\ref{app:repro}).
We write $\mathbf{o}$ for observed frames, available at deployment, and $\mathbf{x}$ for future frames of the demonstration video, available only in training; both are encoded by $\mathcal{E}$ into latents $\mathbf{z}$.
Under diffusion forcing~\citep{diffusionforcing2024}, each latent frame carries an independent noise level $\tau_i \in [0, 1]$: clean observation frames have $\tau_i = 0$, while generative targets draw $\tau_i$ from a shifted logit-normal~\citep{sd3_2024}.

\textbf{Dual-timescale formulation.}
A visual forward world model anticipates future scene evolution $\mathbf{x}_{t+H_f}$ over a foresight horizon $H_f$ conditioned on context.
Existing world-action models couple foresight synchronously to the control rate ($H_f \equiv H_a$)~\citep{cosmospolicy2026,lingbotva2026}, incurring heavy sampling delays and myopic foresight (\S\ref{sec:intro}).
To resolve this, \method{} decouples the foresight horizon ($H_f \approx 3.0$\,s on a slow background clock) from the control rate ($H_a = 0.8$\,s on a fast latent clock).
This decoupling introduces a central challenge: a lookahead latent generated once per $H_f$ is held and reused across ${\approx}H_f/H_a$ consecutive action chunks. 
Consequently, the policy must act against visual foresight whose temporal offset $\Delta$ decays from $H_f$ toward $0$ between refreshes --- a staleness that the training interface must anticipate.

\subsection{Unified Latent World-Action Architecture}
\label{sec:architecture}

\textbf{Three-role training window.}
Training a dual-timescale world-action model requires supervising two concurrent capabilities: generating visual foresight and conditioning actions on that foresight.
To supervise both in a single forward pass, each training sequence provides three distinct visual inputs (Figure~\ref{fig:method}a):
\emph{(i)}~\emph{Observation history} $\mathbf{o}_{\le t}$ ($\tau = 0$) provides clean visual context.
\emph{(ii)}~A noised \emph{future target} $\mathbf{x}_{t+H_f}$ ($\tau \in (0, 1)$) at the full foresight horizon $H_f$ supervises the video backbone's generative forward dynamics.
\emph{(iii)}~A clean \emph{lookahead condition} $\mathbf{x}_{t+u}$ ($\tau = 0$) at a randomized intermediate offset $u \sim \mathcal{U}(0, H_f]$ provides teacher-forced visual guidance for action execution.
Separating the future target from the lookahead condition is essential: while the world model must learn to predict long-horizon transitions at $H_f$, the policy at deployment executes against a held lookahead whose remaining offset $\Delta$ decays over time. Sampling $u \in (0, H_f]$ exposes the policy to this varying offset during training.
The three roles occupy dedicated 3D rotary position embedding (RoPE) temporal slots ($0$, $1$, and $2$).

\textbf{Two-pass causal visual encoding.}
Encoding these three frames into latent tokens requires preventing temporal information leakage during compression.
Because standard 3D causal video VAEs~\citep{wan2025,skyreels2025} aggregate features temporally across frames, encoding all three frames in a single pass would allow future information from $\mathbf{x}_{t+H_f}$ to contaminate the lookahead latent $\mathbf{z}_{t+u}$.
To guarantee strict causal isolation, we encode the inputs in two independent passes of $\mathcal{E}$:
\emph{Pass~1} (generative stream) computes $\mathcal{E}([\mathbf{o}_{\le t}, \mathbf{x}_{t+H_f}]) = [\mathbf{z}_{\le t}, \mathbf{z}_{t+H_f}]$ to provide standard video co-training supervision;
\emph{Pass~2} (policy stream) computes $\mathcal{E}([\mathbf{o}_{\le t}, \mathbf{x}_{t+u}]) = [\mathbf{z}_{\le t}, \mathbf{z}_{t+u}]$, ensuring the lookahead latent depends solely on past context and its own frame.

\begin{wrapfigure}{r}{0.33\textwidth}
\centering
\vspace{-18pt}
\includegraphics[width=0.33\textwidth]{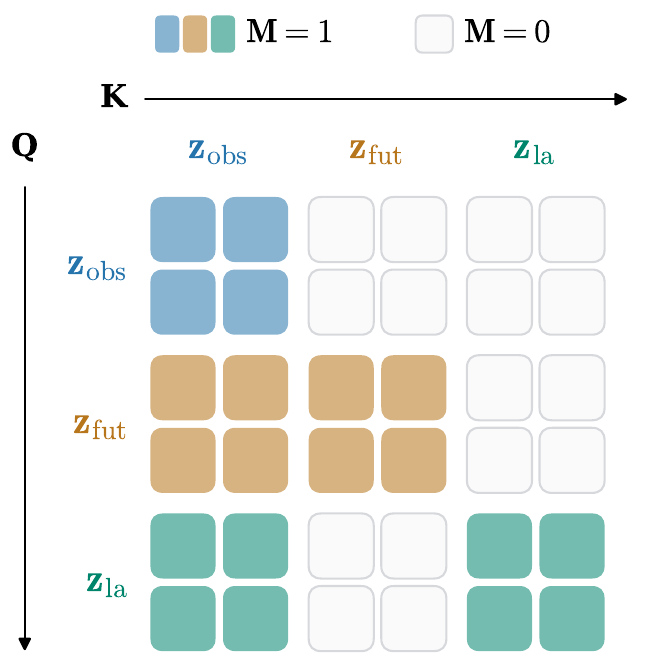}
\vspace{-12pt}
\caption{\raggedright\textbf{Non-interfering attention mask.} 3-class prefix-LM mask $\mathbf{M}$.}
\label{fig:mask}
\vspace{-14pt}
\end{wrapfigure}

\textbf{Non-interfering 3-class attention mask.}
A second leakage path arises inside the transformer: under full self-attention, video generation queries could attend directly to the clean lookahead frame, turning future prediction into a trivial copying shortcut.
To eliminate representation contamination, we design a structured 3-class prefix-LM block mask $\mathbf{M} \in \{0, 1\}^{S \times S}$ (Figure~\ref{fig:mask}) implemented via FlexAttention~\citep{flexattention2024}:
\emph{(i)}~Observation queries attend exclusively to observations ($\mathbf{M}(\mathbf{z}_{\text{obs}}, \mathbf{z}_{\text{obs}}) = 1$).
\emph{(ii)}~Future prediction queries attend to observations and future targets ($\mathbf{M}(\mathbf{z}_{\text{fut}}, \{\mathbf{z}_{\text{obs}}, \mathbf{z}_{\text{fut}}\}) = 1$), but are strictly blocked from lookahead tokens ($\mathbf{M}(\mathbf{z}_{\text{fut}}, \mathbf{z}_{\text{la}}) = 0$).
\emph{(iii)}~Lookahead queries attend to observations and themselves ($\mathbf{M}(\mathbf{z}_{\text{la}}, \{\mathbf{z}_{\text{obs}}, \mathbf{z}_{\text{la}}\}) = 1$).
Because no non-lookahead tokens attend to the lookahead frame ($\mathbf{M}(\cdot, \mathbf{z}_{\text{la}}) = 0$), the backbone representations for observations and future targets remain mathematically identical to standard video co-training, ensuring all policy gains stem strictly from the lookahead conditioning channel.

\subsection{Staleness-Robust Co-Training}
\label{sec:training}

\textbf{Joint flow-matching objective.}
We train the shared video backbone $\theta_{\text{dit}}$ and the action head $\phi_{\text{act}}$ end-to-end via joint conditional flow matching~\citep{flowmatching2023}.
The video objective supervises forward dynamics velocity prediction $\mathbf{v}_\theta$ on the noised future target $\mathbf{z}_{t+H_f}^{(\tau)}$:
\begin{equation}
\mathcal{L}_{\text{video}}(\theta_{\text{dit}}) = \mathbb{E}_{\tau, \boldsymbol{\epsilon}, \mathbf{z}} \left\| \mathbf{v}_\theta\left(\mathbf{z}_{t+H_f}^{(\tau)}, \tau \mid \mathbf{z}_{\le t}, \mathbf{c}\right) - (\boldsymbol{\epsilon} - \mathbf{z}_{t+H_f}) \right\|^2.
\label{eq:loss_video}
\end{equation}
Simultaneously, the action head learns an inverse dynamics model by flow matching on the continuous action chunk $\mathbf{a}_t \in \mathbb{R}^{H_a \times D_a}$, predicting its velocity $\mathbf{u}_\phi$ conditioned on the DiT backbone's multi-layer visual representations $\mathbf{h}$, task instruction $\mathbf{c}$, and lookahead offset $\Delta$:
\begin{equation}
\mathcal{L}_{\text{action}}(\theta_{\text{dit}}, \phi_{\text{act}}) = \mathbb{E}_{\sigma, \boldsymbol{\epsilon}_a, \mathbf{a}} \left\| \mathbf{u}_\phi\left(\mathbf{a}_t^{(\sigma)}, \sigma \mid [\mathbf{h}_{\text{obs}}; \mathbf{h}_{\text{la}}], \mathbf{c}, \Delta\right) - (\boldsymbol{\epsilon}_a - \mathbf{a}_t) \right\|^2,
\label{eq:loss_action}
\end{equation}
where $\sigma$ is the Beta-distributed action flow timestep~\citep{pi0_2024}, and the overall loss is $\mathcal{L} = \mathcal{L}_{\text{video}} + \mathcal{L}_{\text{action}}$.

\textbf{Staleness-robust horizon randomization.}
During asynchronous deployment, a lookahead frame generated at time $t_0$ is held across multiple control cycles, meaning subsequent action chunks at $t_0 + k \cdot H_a$ execute with an aging visual guide whose remaining offset decays toward zero.
To make the policy inherently robust to this staleness without frequent re-generation, we pair the randomized offset sampling $u \sim \mathcal{U}(0, H_f]$ with explicit temporal conditioning:
the action head receives the exact offset $\Delta = u$ via a sinusoidal time embedding~\citep{vaswani2017}.
Exposing the policy to all intermediate offsets during training teaches it to seamlessly follow visual foresight regardless of where the current execution step falls within the refresh cycle.
To retain robust control when foresight is absent or degraded, we apply lookahead token dropout with probability $p=0.1$.

\textbf{Multi-layer visual extraction.}
Rather than extracting features solely from the final DiT block, the action head cross-attends to concatenated representations $[\mathbf{h}_{\text{obs}}; \mathbf{h}_{\text{la}}]$ pooled across four uniformly spaced transformer layers $\{5, 12, 19, 26\}$ (Figure~\ref{fig:method}a).
This multi-layer conditioning combines low-level spatial details from shallow layers with high-level semantic destinations from deep layers, providing a $+0.7\%$ improvement on RoboCasa kitchen (\S\ref{sec:ablations}).

\subsection{Asynchronous Latent-Space Inference}
\label{sec:inference}

\begin{figure}[t]
\centering
\includegraphics[width=\textwidth]{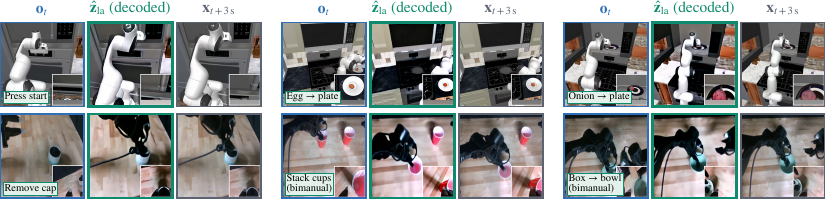}
\caption{\textbf{Imagined lookaheads.} Observation $\mathbf{o}_t$, the \method{} lookahead $\hat{\mathbf{z}}_{\text{la}}$ (decoded for display only), and the real frame $\mathbf{x}_{t+3\,\text{s}}$, in RoboCasa kitchen (top) and real IMETA Y1 rollouts (bottom). The lookahead gets the target layout right while coarsening texture. Wrist view inset; more in Figure~\ref{fig:gallery_full}.}
\label{fig:gallery}
\end{figure}

\textbf{Pure latent-space control path.}
At test time, the model conditions directly on its own generated visual foresight (Figure~\ref{fig:method}b).
Once per horizon $H_f$ (${\sim}3.0$\,s), the video DiT runs an ODE flow sampler for $1$--$10$ steps to generate the lookahead latent $\hat{\mathbf{z}}_{\text{la}}$ from current observation tokens $\mathbf{z}_{\le t}$.
Crucially, $\hat{\mathbf{z}}_{\text{la}}$ is never decoded to raw RGB pixels (Figure~\ref{fig:gallery} decodes some for display): it is retained entirely within the normalized latent space of the causal VAE, directly serving as the slot-$2$ conditioning tokens for subsequent action forward passes.
Eliminating VAE decoding from the control loop removes substantial computational overhead and preserves fine-grained spatial representations.

\textbf{Asynchronous amortization.}
While the lookahead latent is held, the action head decodes subsequent $0.8$\,s action chunks in real time ($48$\,ms per chunk on a single NVIDIA A100 GPU).
Each chunk conditions on the decaying lookahead offset $\Delta = H_f - (t \bmod H_f) \in (0, H_f]$, matching the training distribution of $u$ (\S\ref{sec:training}).
Because a single lookahead frame serves approximately $4$ consecutive action chunks ($H_f / H_a \approx 4$), video sampling overhead is amortized across control cycles (${\sim}17\%$ at $10$ steps, and ${\sim}2\%$ at the $1$-step regime validated in \S\ref{sec:dose}).
By pipelining lookahead generation on a background thread behind active action execution, the lookahead proposer leaves the critical control path entirely, enabling low-latency, closed-loop manipulation.

\section{Experiments}
\label{sec:results}

Our experimental evaluation addresses four questions:
\textbf{(Q1)} How does \method{} compare with state-of-the-art imitation policies and world-action models? (\S\ref{sec:main-results})
\textbf{(Q2)} Does the sparse lookahead transfer to physical single-arm and bimanual robots? (\S\ref{sec:real})
\textbf{(Q3)} Where does the gain come from, and is the lookahead causally load-bearing? (\S\ref{sec:ablations})
\textbf{(Q4)} What does the lookahead cost at deployment, in latency and generative compute? (\S\ref{sec:cost})

\subsection{Experimental Setup}
\label{sec:setup}

\textbf{RoboCasa kitchen.} RoboCasa~\citep{robocasa2024} comprises 24 kitchen manipulation tasks, spanning pick-and-place operations between counters and appliances, door and drawer articulation, knob turning, and button pressing with a Franka Emika Panda arm in procedurally generated scenes. We adopt the evaluation protocol of Cosmos Policy~\citep{cosmospolicy2026}: reporting average success rates across 50 evaluation episodes per task ($n=1200$ total) in five held-out kitchen layouts with unseen object instances (10 episodes per scene). Observations comprise three RGB views (two third-person camera views and one wrist view); Figures~\ref{fig:gallery} and~\ref{fig:gallery_full} show representative tasks alongside model-generated lookahead frames. Following Cosmos Policy, training uses the replay-filtered split of 50 demonstrations per task.

\textbf{LIBERO.} LIBERO~\citep{libero2023} includes four benchmark suites (Spatial, Object, Goal, and Long) containing 10 manipulation tasks each, with 50 demonstrations per task. We evaluate 50 episodes per task (500 episodes per suite, 2000 episodes total) using two RGB camera views (third-person and wrist).

\textbf{Baselines.} Imitation policies: Diffusion Policy~\citep{diffusionpolicy2023}, $\pi_0$~\citep{pi0_2024}, $\pi_0$-fast~\citep{fast2025}, $\pi_{0.5}$~\citep{pi05_2025}, OpenVLA-OFT~\citep{openvlaoft2025}, CogVLA~\citep{cogvla2025}, and GR00T-N1/N1.5~\citep{gr00t2025} with data augmentation (+DreamGen, +DUST, +HAMLET). World-action models: UVA~\citep{uva2025}, UWM~\citep{uwm2025}, Video Policy~\citep{turningvideo2026}, FLARE~\citep{flare2025}, Cosmos Policy~\citep{cosmospolicy2026}, LingBot-VA~\citep{lingbotva2026}, Fast-WAM~\citep{fastwam2026}, Motus, Enfold-Flash, and DiT4DiT. Baseline numbers are from the original publications; all internal comparisons share datasets, backbones, and evaluation pipelines.

\textbf{Training details.} All \method{} variants and internal baselines are implemented in StarVLA~\citep{starvla2026} and initialize from the pretrained SkyReels-V2-DF-1.3B backbone~\citep{skyreels2025} and train strictly on demonstration data without online rollouts, specialized data curation, or auxiliary labels beyond RGB images and robot actions. All models are trained on $4\times$ NVIDIA H200 GPUs. Benefiting from effective pretraining and sparse lookahead conditioning, \method{} converges rapidly: it requires only 10k training steps on RoboCasa kitchen and 15k steps on LIBERO to reach top performance, substantially faster than typical baseline training horizons (e.g., 60k--100k+ steps). We evaluate EMA checkpoints (details in Appendix~\ref{app:repro}). In simulation, the client pauses the simulator while each lookahead is generated (asynchronous execution with zero delay; \S\ref{sec:cost} tests a pipelined schedule); on the real robot, \method{} runs fully asynchronously on a single machine.

\begin{table}[t]
\centering\scriptsize
\renewcommand{\arraystretch}{0.95}
\setlength{\tabcolsep}{2.5pt}
\begin{minipage}[t]{0.44\textwidth}
\centering
\caption{\textbf{RoboCasa kitchen.} 24-task average SR, held-out scenes; baselines as
reported~\citep{gr00t2025,pi0_2024,uwm2025,turningvideo2026,flare2025,cosmospolicy2026}.
Blocks: VLA policies, WAMs.}
\label{tab:robocasa}
\vspace{2pt}
\begin{tabular}[t]{lcc}
\toprule
Model & SR (\%) & Demos/task \\
\midrule
GR00T-N1            & 49.6 & 300 \\
GR00T-N1 + DreamGen & 57.6 & 300 \\
GR00T-N1 + DUST     & 58.5 & 300 \\
$\pi_0$             & 62.5 & 300 \\
GR00T-N1.5          & 64.1 & 300 \\
GR00T-N1.5 + HAMLET & 66.4 & 300 \\
\midrule
UVA                 & 50.0 & 300 \\
UWM                 & 60.8 & 300 \\
Video Policy        & 66.0 & 300 \\
FLARE               & 66.4 & 300 \\
Cosmos Policy       & \underline{67.1} & 50 \\
\midrule
\textbf{\method{} (ours)} & \textbf{72.2} & 50 \\
\bottomrule
\end{tabular}
\end{minipage}\hfill
\begin{minipage}[t]{0.54\textwidth}
\centering
\caption{\textbf{LIBERO.} Per-suite SR as
reported~\citep{diffusionpolicy2023,fast2025,pi05_2025,openvlaoft2025,cogvla2025,cosmospolicy2026,lingbotva2026,fastwam2026,enfold2026};
blocks as in Table~\ref{tab:robocasa}; best bold, second underlined. Sizes and latencies: Figure~\ref{fig:latency}.}
\label{tab:libero}
\vspace{2pt}
\begin{tabular}[t]{lccccc}
\toprule
Method & Spatial & Object & Goal & Long & Avg \\
\midrule
Diffusion Policy & 78.3 & 92.5 & 68.3 & 50.5 & 72.4 \\
$\pi_0$-fast     & 96.4 & 96.8 & 88.6 & 60.2 & 85.5 \\
$\pi_{0.5}$      & \underline{98.8} & 98.2 & 98.0 & 92.4 & 96.9 \\
OpenVLA-OFT      & 97.6 & 98.4 & 97.9 & 94.5 & 97.1 \\
CogVLA           & 98.6 & 98.8 & 96.6 & 95.4 & 97.4 \\
\midrule
Motus            & 96.8 & \underline{99.8} & 96.6 & 97.6 & 97.7 \\
DiT4DiT          & 98.4 & 99.6 & \underline{98.6} & 97.6 & \underline{98.6} \\
Fast-WAM         & 98.2 & \textbf{100.0} & 97.0 & 95.2 & 97.6 \\
Enfold-Flash     & 97.0 & \underline{99.8} & 96.6 & 96.6 & 97.5 \\
Cosmos Policy    & 98.1 & \textbf{100.0} & 98.2 & 97.6 & 98.5 \\
LingBot-VA       & 98.5 & 99.6 & 97.2 & \textbf{98.5} & 98.5 \\
\midrule
\textbf{\method{} (ours)} & \textbf{99.4} & \textbf{100.0} & \textbf{99.0} & \underline{97.8} & \textbf{99.0} \\
\bottomrule
\end{tabular}
\end{minipage}
\end{table}

\subsection{(Q1) Comparison with State-of-the-Art Policies}
\label{sec:main-results}

Tables~\ref{tab:robocasa} and~\ref{tab:libero} evaluate \method{} against top-performing imitation policies and world-action models. On the 24-task RoboCasa kitchen benchmark, \method{} achieves an average success rate of $72.2\%$ with 50 demonstrations per task, outperforming Cosmos Policy by $+5.1\%$ under matched demonstration data and exceeding all baselines trained with $6\times$ more demonstration episodes (300 vs.\ 50). On LIBERO, \method{} attains $99.0\%$ average success across the four suites.

\begin{figure}[t]
\centering
\includegraphics[width=\textwidth]{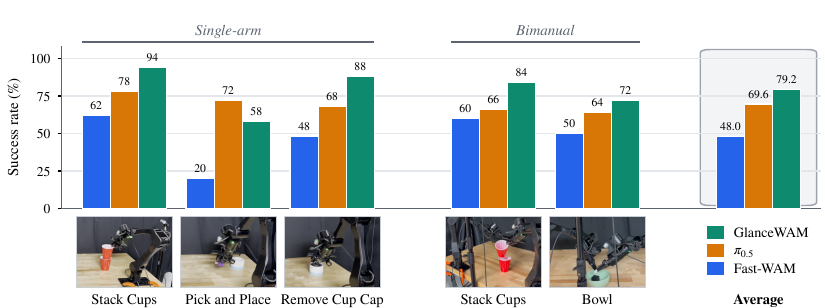}
\caption{\textbf{Real-robot evaluation on IMETA Y1.} Success of \method{}, $\pi_{0.5}$, and Fast-WAM over 50 trials per task; thumbnails show one \method{} rollout per task.}
\label{fig:real}
\end{figure}

\subsection{(Q2) Real-Robot Evaluation}
\label{sec:real}

We deploy \method{} on the IMETA Y1 platform against $\pi_{0.5}$~\citep{pi05_2025} and Fast-WAM~\citep{fastwam2026} in two configurations (Figure~\ref{fig:real}): a single arm (side and wrist cameras) on \emph{stack cups}, \emph{pick and place}, and \emph{remove cup cap}, and a dual-arm setup (top camera plus one wrist camera per arm) on bimanual \emph{stack cups} and \emph{bowl} (pick up the bowl and put the box into it). All policies are fine-tuned on identical 50 demonstrations per task, deployed with LeRobot~\citep{lerobot2026} and real-time chunking (RTC)~\citep{rtc2025}, and evaluated over 50 physical trials per task under randomized initializations. \method{} achieves $79.2\%$ average success vs.\ $69.6\%$ for $\pi_{0.5}$ and $48.0\%$ for Fast-WAM, outperforming $\pi_{0.5}$ on macro-reconfigurations: single-arm \emph{stack cups} ($94\%$ vs.\ $78\%$) and \emph{remove cup cap} ($88\%$ vs.\ $68\%$), and both bimanual tasks ($84\%$ vs.\ $66\%$ on \emph{stack cups}, $72\%$ vs.\ $64\%$ on \emph{bowl}). This holds although $\pi_{0.5}$ is pretrained on large-scale robot data, whereas \method{} starts from an internet video prior and sees robot data only in the 50 demonstrations. Fast-WAM, which keeps video modeling at training time only, trails both on every task. \method{} trails $\pi_{0.5}$ only on \emph{pick and place} ($58\%$ vs.\ $72\%$), where the $3{\times}3$\,cm cube spans just 1--2 latent tokens and millimeter grasp alignment falls to local visual feedback (Appendix~\ref{app:real}).

\subsection{(Q3) Where Does the Gain Come From?}
\label{sec:ablations}

\begin{wraptable}{r}{0.43\textwidth}
\centering\scriptsize
\vspace{-11pt}
\caption{\textbf{Internal comparison on RoboCasa kitchen} (24-task SR, held-out scenes, demos only; shared backbone, data, protocol). \textbf{(a)}~Components. \textbf{(b)}~Lookahead sampler steps, single-layer checkpoint, paired re-evaluation ($1{,}200$ episodes/point). $^\dagger$Correct lookahead, $\Delta$ fixed to one frame; paired sweep, 1{,}200 episodes (baseline $71.3$, zeroed $62.7$).}
\label{tab:main}\label{tab:gensteps}
\vspace{2pt}
\setlength{\tabcolsep}{2.5pt}
\renewcommand{\arraystretch}{0.95}
\begin{tabular}{lc}
\toprule
(a) System & SR (\%) \\
\midrule
Cosmos Policy (external anchor) & 67.1 \\
\midrule
plain co-training (no lookahead) & 64.4 \\
\method{}, single-layer lookahead & 71.5 \\
\quad same checkpoint, lookahead zeroed & 61.6 \\
\quad same checkpoint, offset $\Delta$ pinned$^\dagger$ & 61.8 \\
\quad retrained without mask (bidirectional) & 47.0 \\
\textbf{\method{}, multi-layer (final)} & \textbf{72.2} \\
\bottomrule
\end{tabular}\\[4pt]
\setlength{\tabcolsep}{3.5pt}
\begin{tabular}{lccccc}
\toprule
(b) Euler steps & 1 & 2 & 5 & 10 & 30 \\
\midrule
SR (\%) & 71.2 & 71.7 & 71.4 & 71.5 & 69.8 \\
\bottomrule
\end{tabular}
\vspace{-8pt}
\end{wraptable}

\textbf{Component isolation.} Table~\ref{tab:main}a systematically isolates the contribution of each design component. Standard video--action co-training without lookahead conditioning (using the same backbone, data, and compute budget) reaches $64.4\%$ success, demonstrating that representation learning alone at a short action horizon does not bridge the performance gap to Cosmos Policy ($67.1\%$). Introducing the lookahead conditioning channel via single-layer extraction yields a $+7.1\%$ improvement ($71.5\%$), and extracting lookahead features across four transformer layers (\S\ref{sec:training}) provides an additional $+0.7\%$, reaching $72.2\%$. Replicated lookahead runs land at $71.4\%$ and $71.2\%$, while lookahead-free co-training plateaus near $64\%$ even at 60k steps. Retraining the single-layer model without the non-interfering mask (bidirectional attention, otherwise identical recipe) collapses success to $47.0\%$.

\textbf{The lookahead is causally load-bearing.}\label{sec:causal}
To test whether lookahead features are causally responsible for action decisions rather than merely present, we apply an evaluation-time intervention that sets all lookahead tokens to zero on the trained checkpoint. Success drops from $71.5\%$ to $61.6\%$ (Table~\ref{tab:main}a), below the baseline trained without lookaheads ($64.4\%$): the policy has learned to rely on the lookahead. Keeping the correct lookahead but pinning its offset $\Delta$ to one frame falls to the same floor ($61.8\%$): the head also reads the offset.

\begin{figure}[t]
\centering
\begin{minipage}[t]{0.48\textwidth}
\centering
\vspace{0pt}
\includegraphics[width=\linewidth]{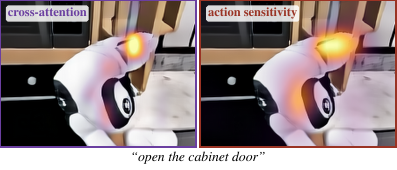}
\vspace{-14pt}
\caption{\textbf{The action head reads the lookahead.} Value-weighted cross-attention and action sensitivity on a decoded lookahead $\hat{\mathbf{z}}_{\text{la}}$; more in Figure~\ref{fig:attn_app}.}
\label{fig:attn}
\end{minipage}\hfill
\begin{minipage}[t]{0.48\textwidth}
\centering
\vspace{0pt}
\begin{tikzpicture}
\begin{axis}[
    width=\linewidth,
    height=4.15cm,
    grid=both,
    grid style={dotted, gray!35},
    xlabel={prediction horizon $H$},
    ylabel={RoboCasa SR (\%)},
    xlabel style={font=\tiny, yshift=4pt},
    ylabel style={font=\tiny, yshift=-5pt},
    tick label style={font=\tiny},
    xmin=0.5, xmax=4.1,
    ymin=61.0, ymax=75.0,
    xtick={0.8, 1.5, 2.2, 3.0, 3.8},
    xticklabels={0.8\,s, 1.5\,s, 2.2\,s, 3.0\,s, 3.8\,s},
    ytick={62, 66, 70, 74},
]

\addplot[
    color=cgoal,
    mark=*,
    mark size=2.0pt,
    line width=1.1pt,
    mark options={fill=cgoal!40, draw=cgoal, line width=0.7pt}
] coordinates {
    (1.4, 66.3)
    (2.2, 70.1)
    (3.0, 71.6)
    (3.8, 71.1)
};

\node[font=\tiny\bfseries, text=cgoal!90!black, anchor=south] at (axis cs:3.0, 71.8) {71.6\%};
\node[font=\tiny, text=cgoal!90!black, anchor=west] at (axis cs:0.6, 72.4) {\method{} ($H_f$)};

\addplot[
    color=cobs,
    dashed,
    mark=square*,
    mark size=1.8pt,
    line width=0.9pt,
    mark options={fill=cobs!40, draw=cobs, line width=0.7pt}
] coordinates {
    (0.8, 63.1)
    (1.6, 65.7)
    (2.4, 64.3)
    (3.0, 63.7)
    (3.8, 65.3)
};

\node[font=\tiny, text=cobs!90!black, anchor=east] at (axis cs:4.02, 67.3) {plain co-training ($H_v$)};

\end{axis}
\end{tikzpicture}
\vspace{-14pt}
\caption{\textbf{Prediction horizon.} RoboCasa SR vs.\ lookahead $H_f$ and plain co-training target $H_v$.}
\label{fig:horizon}
\end{minipage}
\vspace{-6pt}
\end{figure}

\textbf{The action head reads the lookahead.}\label{sec:attn}
We probe the action head's cross-attention over $[\mathbf{h}_{\text{obs}};\mathbf{h}_{\text{la}}]$ on 12 demonstration contexts (single-layer checkpoint of Table~\ref{tab:main}a). Lookahead tokens supply $40\%$ of the action queries' cross-attention output (value-weighted share; $34\%$ of the raw softmax mass). Two independent measurements agree on where the lookahead is read (Figure~\ref{fig:attn}): the value-weighted attention of each lookahead token, and the causal action shift when a $2{\times}2$-token lookahead patch is overwritten with the corresponding observation tokens. Both peak on the interaction the lookahead depicts (the gripper at the cabinet-door edge; the mug at the dispenser in Figure~\ref{fig:attn_app}), correlate spatially within a context ($r = 0.55$, mean over the 12 contexts), and split across cameras alike ($0.28/0.32/0.41$ vs.\ $0.27/0.29/0.43$ for third-person L/R and wrist).

\textbf{How far ahead should the prediction target be?}
Figure~\ref{fig:horizon} examines the effect of prediction horizons across both paradigms. For plain co-training (representation shaping), predicting future targets at $H_v = 1.6$--$2.4$\,s outperforms the action-chunk horizon of $0.8$\,s ($63.1\% \to 65.7\%$): predicting beyond immediate transitions enriches the representation. For \method{} (visual foresight), success rises from $66.3\%$ at $H_f = 1.4$\,s to a peak of $71.6\%$ at $3.0$\,s (60 frames) and plateaus at $3.8$\,s ($71.1\%$): foresight helps most when it anticipates distal subgoals (${\sim}3$\,s) rather than short-horizon transitions.

\subsection{(Q4) Deployment Cost}
\label{sec:cost}

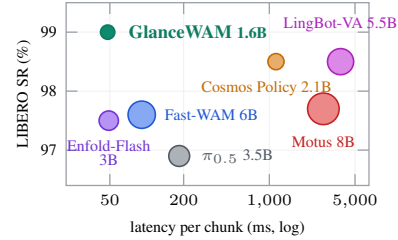
\begin{wrapfigure}{r}{0.40\textwidth}
\centering
\vspace{-12pt}
\definecolor{cfastwam}{HTML}{2563EB}  %
\definecolor{ccosmos}{HTML}{D97706}   %
\definecolor{clingbot}{HTML}{C026D3}  %
\definecolor{cmotus}{HTML}{DC2626}    %
\definecolor{cenfold}{HTML}{7C3AED}   %
\begin{tikzpicture}
\begin{axis}[
  width=\linewidth,
  height=4.0cm,
  xmode=log,
  log ticks with fixed point,
  xmin=22, xmax=6500,
  ymin=96.4, ymax=99.5,
  xlabel={latency per chunk (ms, log)},
  ylabel={LIBERO SR (\%)},
  xlabel near ticks,
  ylabel near ticks,
  label style={font=\tiny}, xlabel style={yshift=2pt}, ylabel style={yshift=-3pt},
  tick label style={font=\tiny},
  xtick={50,200,1000,5000},
  ytick={97,98,99},
  yticklabels={97,98,99},
  axis line style={cgray!60, line width=0.5pt},
  tick style={cgray!60, line width=0.5pt},
  ymajorgrids,
  grid style={cgray!18, line width=0.35pt},
  xmajorgrids,
  grid style={cgray!10, line width=0.25pt},
  clip=false
]
\addplot[only marks, mark=*, mark size=5.14pt, draw=cfastwam!90!black, fill=cfastwam!55, line width=0.6pt]
  coordinates {(91,97.6)};
\node[font=\tiny, text=cfastwam!90!black, anchor=west] at (axis cs:118,97.6) {Fast-WAM 6B};
\addplot[only marks, mark=*, mark size=3.04pt, draw=ccosmos!90!black, fill=ccosmos!60, line width=0.6pt]
  coordinates {(1133,98.5)};
\node[font=\tiny, text=ccosmos!90!black, anchor=north] at (axis cs:950,98.33) {Cosmos Policy 2.1B};
\addplot[only marks, mark=*, mark size=4.92pt, draw=clingbot!90!black, fill=clingbot!55, line width=0.6pt]
  coordinates {(3812,98.5)};
\node[font=\tiny, text=clingbot!90!black, anchor=south] at (axis cs:3812,98.86) {LingBot-VA 5.5B};
\addplot[only marks, mark=*, mark size=5.94pt, draw=cmotus!90!black, fill=cmotus!55, line width=0.6pt]
  coordinates {(2759,97.7)};
\node[font=\tiny, text=cmotus!90!black, anchor=north] at (axis cs:2759,97.40) {Motus 8B};
\addplot[only marks, mark=*, mark size=3.64pt, draw=cenfold!90!black, fill=cenfold!55, line width=0.6pt]
  coordinates {(49,97.5)};
\node[font=\tiny, text=cenfold!90!black, anchor=north, align=center] at (axis cs:49,97.32) {Enfold-Flash\\[-2pt]3B};
\addplot[only marks, mark=*, mark size=3.93pt, draw=cgray!90!black, fill=cgray!50, line width=0.6pt]
  coordinates {(184,96.9)};
\node[font=\tiny, text=cgray!90!black, anchor=west] at (axis cs:235,96.9) {$\pi_{0.5}$ 3.5B};
\addplot[only marks, mark=*, mark size=2.65pt, draw=cgoal!90!black, fill=cgoal, line width=0.7pt]
  coordinates {(48,99.0)};
\node[font=\scriptsize\bfseries, text=cgoal!90!black, anchor=west] at (axis cs:60,99.0) {\method{} {\tiny 1.6B}};
\end{axis}
\end{tikzpicture}
\vspace{-14pt}
\caption{\textbf{Latency vs.\ success on LIBERO.} Per-chunk latency ($B{=}1$); marker area: model size.}
\label{fig:latency}
\vspace{-10pt}
\end{wrapfigure}

\textbf{Real-time latent-space decoding.} With one clean DiT pass plus the action head and no VAE decoding (\S\ref{sec:inference}), the control path takes 48\,ms per 16-action chunk on LIBERO at $B=1$ on an NVIDIA A100 40GB GPU (breakdown in Appendix~\ref{app:latency}). Figure~\ref{fig:latency} illustrates the latency--performance landscape across world-action models: synchronous architectures that reach comparable success rates require $24$--$80\times$ higher inference latency ($1133$--$3812$\,ms), while accelerated alternatives such as Fast-WAM ($91.5$\,ms compiled) and Enfold-Flash ($49$\,ms) exhibit lower task success ($97.5$--$97.6\%$) while maintaining $2$--$4\times$ larger parameter counts (3--6B vs.\ 1.6B). Lookahead generation ($478.6$\,ms per refresh at 10 steps, ${\approx}145$\,ms at 1 step; Appendix~\ref{app:latency}) runs in the background and never enters the control loop.

A pipelined variant that refreshes the lookahead every chunk changes success by $-0.5\%$ relative to the blocking evaluation ($p=0.72$), provided lookaheads are refreshed at the scheduled cadence (paired re-evaluation, 1{,}200 episodes).

\textbf{One sampler step suffices.}\label{sec:dose}
Sweeping the lookahead sampler over $1$--$30$ Euler steps (a $30\times$ compute span; Table~\ref{tab:gensteps}b) moves success by at most $1.9\%$ ($71.2\%$ at 1 step vs.\ $71.5\%$ at 10), although image sharpness visibly differs (Figure~\ref{fig:goaldemos}, Appendix~\ref{app:qualitative}). A 1-step sampler thus cuts proposer compute $10\times$ at no measurable cost: the policy reads low-frequency spatial layout.

\section{Conclusion}
\label{sec:conclusion}

Synchronous world-action models tie visual prediction to the action-chunk rate, which makes their foresight both slow and myopic. \method{} decouples the two on one video DiT: it imagines a single latent lookahead frame seconds ahead, off the control path, and decodes actions against it in 48\,ms, reaching $72.2\%$ on RoboCasa kitchen, $99.0\%$ on LIBERO, and higher real-robot success than $\pi_{0.5}$ and Fast-WAM. Sparse, asynchronous foresight lets world-action models keep test-time imagination without paying for it in the control loop.

\subsection*{Reproducibility statement}
Source code is publicly available at \url{https://github.com/linhanwang/GlanceWAM}, comprising training and simulation evaluation code built on StarVLA~\citep{starvla2026} and real-robot deployment code built on LeRobot~\citep{lerobot2026} with RTC~\citep{rtc2025}; real-robot rollout videos for all five tasks are on the project page (\url{https://linhanwang.github.io/glancewam/}).
Simulation experiments use the public RoboCasa and LIBERO benchmarks with their standard demonstration splits and evaluation protocols (\S\ref{sec:setup}).
Model architecture, hyperparameters, and training configuration are given in Appendix~\ref{app:repro}, latency measurement in Appendix~\ref{app:latency}, and the physical robot setup in Appendix~\ref{app:real}.

\subsection*{AI use statement}
In this work, we used generative AI tools to assist in implementing methods (writing and editing software code).
We have not used generative AI tools to generate synthetic data, develop theoretical models or conceptual frameworks, formulate or prove mathematical claims, propose or refine hypotheses, design research methodology or experiments, clean or reformat datasets, or interpret results; translation and qualitative data analysis are not applicable to this work.
Additionally, we used generative AI tools to edit the paper to improve readability.
We have reviewed all AI-assisted work: the authors carefully read the full paper and all AI-assisted code, and verified the code through rigorous testing.
We take responsibility for the final content of this work, including text, claims or artifacts produced with the aid of generative AI.

\bibliography{references}
\bibliographystyle{iclr2027_conference}

\appendix

\section{Concurrent work}
\label{app:concurrent}

Two concurrent world-action models appeared while this work was in
preparation (Table~\ref{tab:concurrent}). DeVA~\citep{deva2026} (July 2026)
pairs a Cosmos-Predict2 video expert with a GR00T-style action DiT through a
multi-level feature bridge, supervises the video features with affordance and
depth decoders (labels from contact annotations and a pretrained depth model),
and denoises video and action jointly in one synchronous pass.
Flex-$\pi$~\citep{flexpi2026} (August 2026) is a 6B WAM that jointly denoises
RGB, 3D-pointmap, and DINOv3 futures with actions, and uses per-stream dropout
to fall back to action-only inference at deployment.

\begin{table}[h]
\centering\small
\setlength{\tabcolsep}{5pt}
\caption{\textbf{Comparison with concurrent world-action models.}
Benchmark numbers as reported by each paper under its own evaluation protocol.
The DeVA ablation row removes its affordance and depth decoders, leaving RGB
demonstrations only, as in \method{}. Flex-$\pi$ does not report RoboCasa
kitchen; its two LIBERO rows are the reported FLEX-$\pi^*$ deployment modes.}
\label{tab:concurrent}
\vspace{2pt}
\resizebox{\textwidth}{!}{%
\begin{tabular}{lcccc}
\toprule
 & RoboCasa & LIBERO & Auxiliary supervision & World model \\
 & kitchen & (avg) & beyond RGB demos & at test time \\
\midrule
DeVA~\citep{deva2026} & 72.0 & 99.0 &
  \begin{tabular}[t]{@{}c@{}}affordance + depth decoders\\(contact labels, depth model)\end{tabular} &
  \begin{tabular}[t]{@{}c@{}}joint denoising,\\every chunk\end{tabular} \\
\addlinespace[2pt]
\quad w/o guidance (their ablation) & 66.0 & --- & none & (same) \\
\addlinespace[2pt]
Flex-$\pi$~\citep{flexpi2026}, action-only & --- & 98.7 &
  \begin{tabular}[t]{@{}c@{}}3D pointmaps + DINOv3\\semantic futures\end{tabular} &
  dropped (fast path) \\
\addlinespace[2pt]
Flex-$\pi$, full joint & --- & 99.2 & (same) &
  \begin{tabular}[t]{@{}c@{}}joint denoising,\\every chunk (slow path)\end{tabular} \\
\addlinespace[2pt]
\textbf{\method{} (ours)} & \textbf{72.2} & 99.0 & none &
  async lookahead frame, $\sim$3\,s \\
\bottomrule
\end{tabular}}
\end{table}

\section{Latency measurement}
\label{app:latency}

\textbf{Protocol.} All \method{} latencies are measured at $B{=}1$ on a single
NVIDIA A100 (SXM4 40\,GB, bf16), $n{=}100$ timed calls after 15 warmup calls,
CUDA-synchronized wall clock, on the released LIBERO checkpoint (multi-layer,
$99.0\%$ in Table~\ref{tab:libero}). Inputs match the evaluation client: two
$256^2$ views (agent + wrist) stitched to $224{\times}448$, one observation
frame, and a 16-action chunk from the 4-step flow-matching head. Since the
lookahead proposer runs asynchronously (\S\ref{sec:cost}), we report the
hold-phase action path: \texttt{predict\_action} with the current lookahead
latent already in hand.

\textbf{Deployment inference path.} The timed path uses the same optimizations
as our serving stack: the video DiT is truncated after the last feature layer
(blocks $1..27$ of $30$ for layers $\{5,12,19,26\}$; the velocity head is not
needed for actions); the VAE encoder, truncated DiT, and action head are
compiled with CUDA graphs (\texttt{torch.compile}, reduce-overhead);
preprocessing runs on GPU; and the text embedding, lookahead latent, and
attention block mask are cached. This measures $47.95$\,ms per chunk (std
$0.39$\,ms), which we quote as 48\,ms; serving adds a websocket round trip of
$\approx$1--2\,ms. The unoptimized path (full 30-block forward, per-block
compilation) measures $108.3$\,ms (std $1.15$\,ms).

\textbf{Lookahead refresh.} A refresh call (generate the lookahead latent,
then act on it) measures 478.6\,ms at the trained 10-step sampler budget on the
unoptimized path, i.e.\ $\approx$37\,ms per Euler step on top of the 108.3\,ms
hold path; at the 1-step budget (\S\ref{sec:dose}) it is $\approx$145\,ms. In
the asynchronous design this cost never blocks the control loop: the proposer
runs in the background once per $H_f$ (every 3--4 chunks) while the head keeps
acting on the held lookahead, so the per-chunk control cost stays at 48\,ms.

\textbf{Same-hardware baselines.} Baseline latencies in
Figure~\ref{fig:latency} are measured on the same GPU class (A100 40\,GB). We
re-benchmarked Fast-WAM (6B MoT, 32-action chunk, 10-step action head) on our
A100: its compiled path (\texttt{torch.compile} reduce-overhead with CUDA
graphs) takes 91.5\,ms per chunk, matching the published 91\,ms (634\,ms in
eager mode; Enfold~\citep{enfold2026} reports 493\,ms). Neither system
generates video in the per-chunk action pass; \method{} is ${\approx}2\times$
faster than Fast-WAM ($48$ vs.\ $91.5$\,ms) with higher LIBERO success
($99.0\%$ vs.\ $97.6\%$) and fewer parameters ($1.6$B vs.\ $6$B).

\section{Additional figures}
\label{app:qualitative}

\begin{figure}[t]
\centering
\includegraphics[width=\textwidth]{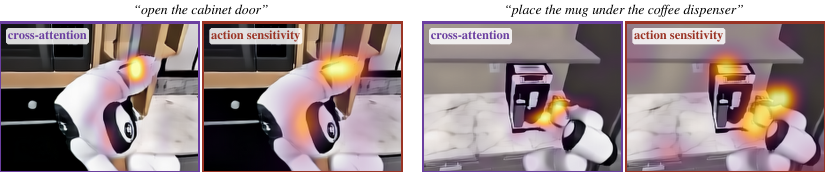}
\caption{\textbf{Where the action head reads the lookahead} (extends Figure~\ref{fig:attn}). Two probed contexts, third-person view of the decoded lookahead $\hat{\mathbf{z}}_{\text{la}}$. \emph{Cross-attention:} value-weighted contribution of each lookahead token to the action queries' cross-attention output, i.e.\ the norm of its attention-weighted value after the output projection. The value-weighted share in \S\ref{sec:attn} is $\|\text{lookahead part}\| / (\|\text{obs part}\| + \|\text{lookahead part}\|)$ of the attention output. \emph{Action sensitivity:} action shift when a $2{\times}2$-token lookahead patch is overwritten with the observation tokens at the same location. Each map is normalized independently.}
\label{fig:attn_app}
\end{figure}

\begin{figure}[t]
\centering
\includegraphics[width=\textwidth]{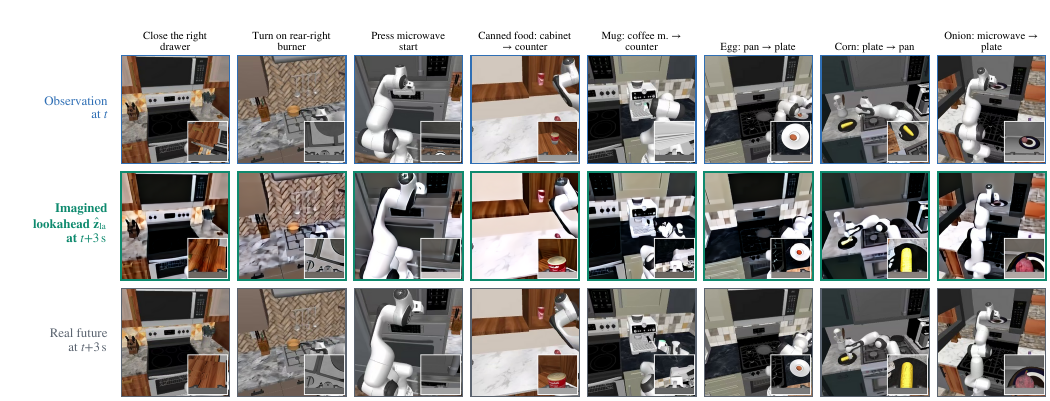}
\caption{\textbf{Lookahead frames on eight kitchen tasks} (extends Figure~\ref{fig:gallery}). For eight RoboCasa kitchen tasks (articulation, knobs, buttons, pick-and-place): the observation at $t$, the \method{} lookahead, and the environment frame at $t+3$\,s. Lookaheads get the target layout right (closed doors, objects on target surfaces) while coarsening texture; they are decoded to RGB only for display. One third-person view per panel, wrist view inset.}
\label{fig:gallery_full}
\end{figure}

\begin{figure*}[t]
\centering
\includegraphics[width=0.98\textwidth]{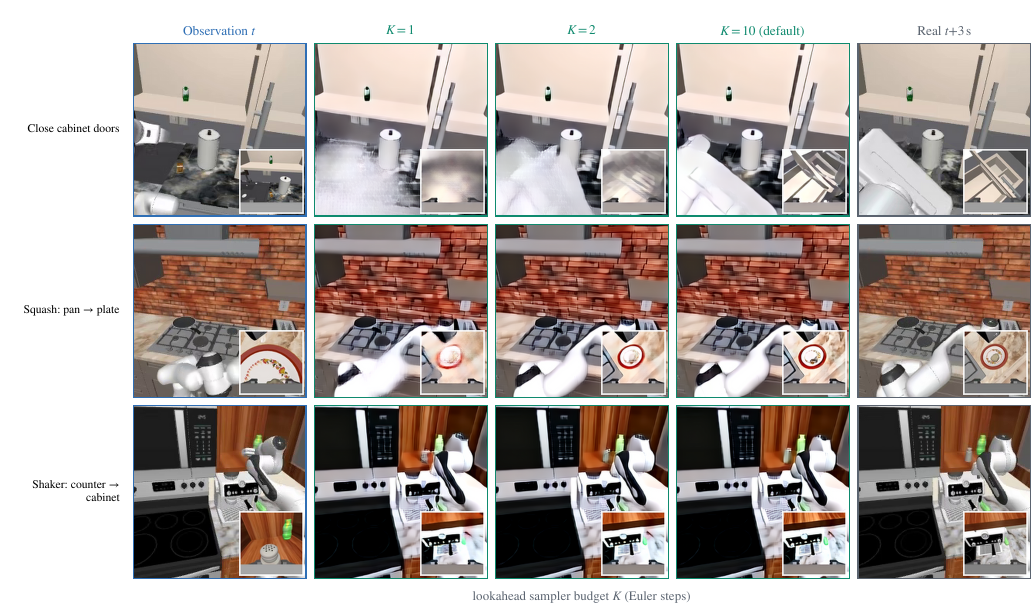}
\caption{\textbf{Generated lookaheads vs.\ sampler budget $K$} (RoboCasa
kitchen demonstration contexts, matched seeds per row; last column: the
demonstration frame at $t{+}H_f$). Fidelity improves from $K{=}1$ to $K{=}10$
(the one-step lookahead smears the arm and the manipulated object, most visibly
in the wrist insets), yet success barely changes over $K = 1$--$30$
(Table~\ref{tab:gensteps}b). Layout as in Figure~\ref{fig:gallery_full}.}
\label{fig:goaldemos}
\end{figure*}

Figure~\ref{fig:attn_app} extends the attention probe of \S\ref{sec:attn},
Figure~\ref{fig:gallery_full} extends the lookahead gallery of
Figure~\ref{fig:gallery}, and Figure~\ref{fig:goaldemos} shows lookaheads across
the sampler budgets of Table~\ref{tab:gensteps}b: fidelity improves with
compute, success does not.

\section{Physical robot deployment details and rollouts}
\label{app:real}

\paragraph{Setup.}
We use the IMETA Y1 platform in two configurations. The single-arm setup is one
6-DoF Y1 arm with a parallel-jaw gripper, observed by a fixed side camera and a
wrist camera. The bimanual setup uses two Y1 arms, observed by a fixed top
camera and one wrist camera per arm. All cameras record $640{\times}480$ RGB,
cropped to the workspace and resized to $224{\times}224$ for the backbone.
Actions are absolute joint positions plus gripper position for each arm (7-D
single-arm, 14-D bimanual), recorded and executed at $20$\,Hz. \method{} runs
fully asynchronously on a single machine, with lookahead generation hidden
behind action execution.
The deployment stack is built on LeRobot~\citep{lerobot2026}, and all three
policies execute action chunks with real-time chunking (RTC)~\citep{rtc2025}.

\paragraph{Tasks and protocol.}
Three single-arm and two bimanual tasks:
\begin{itemize}
\item \emph{Stack cups} (single-arm): grasp the right cup by the rim and nest it in the left cup without toppling either.
\item \emph{Pick and place} (single-arm): pick up a $3{\times}3{\times}3$\,cm foam cube and drop it into an open cup.
\item \emph{Remove cup cap} (single-arm): pull the fitted lid off a closed container and place it on the table.
\item \emph{Stack cups} (bimanual): stack the two cups using both arms without toppling them.
\item \emph{Bowl} (bimanual): one arm lifts the bowl and the other places a box into it.
\end{itemize}
Each task has 50 teleoperated demonstrations, and $\pi_{0.5}$, Fast-WAM, and
\method{} are fine-tuned on the same data. %
Each policy is evaluated over 50 trials per task (250 in total), with initial
object poses perturbed within $\pm 8$\,cm and $\pm 30^\circ$. A trial fails if
an object is dropped, a container topples, or the task takes longer than
30\,s.

\begin{figure*}[t]
\centering
\includegraphics[width=0.98\textwidth]{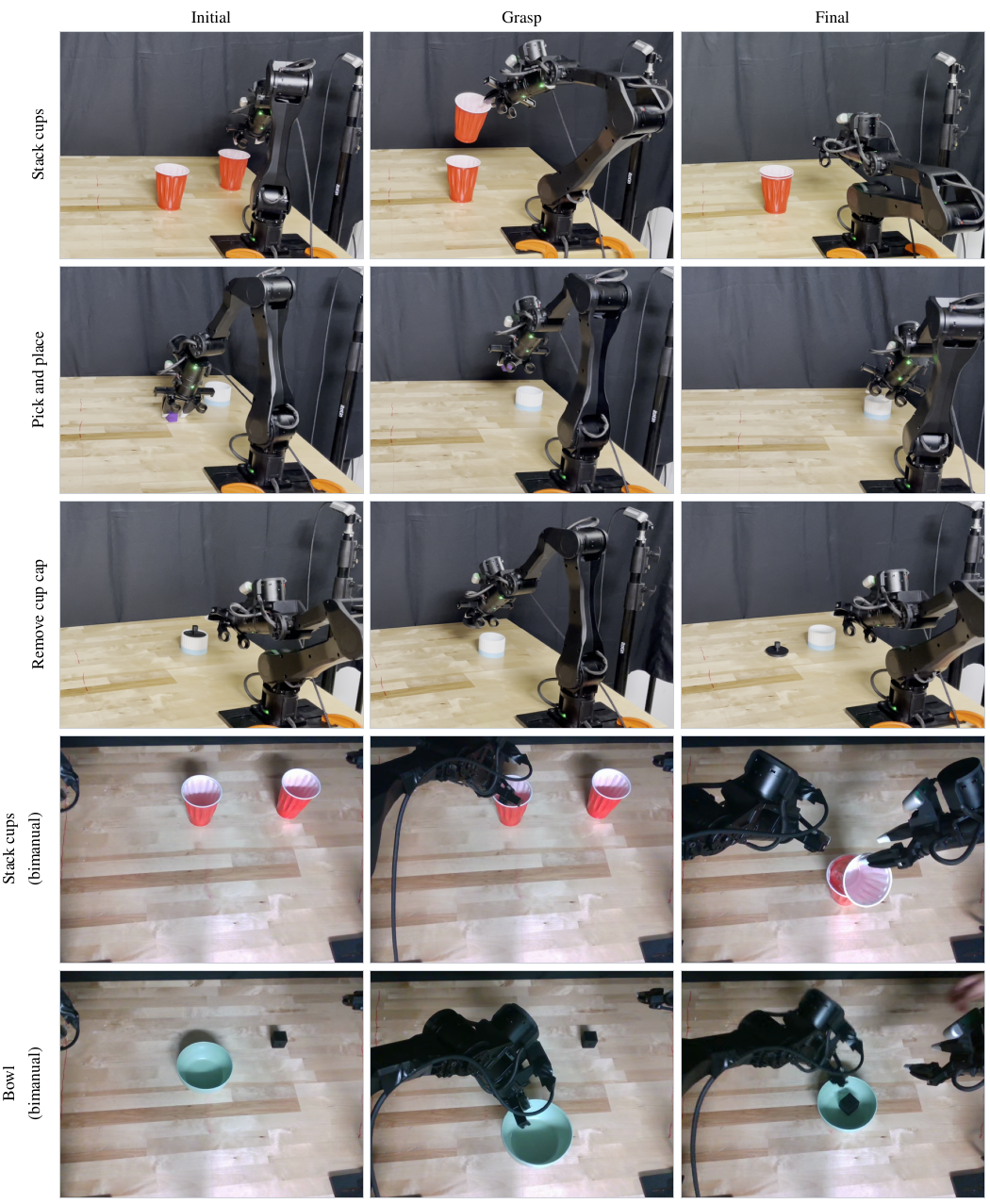}
\caption{\textbf{Real-robot rollout keyframes.} Initial, grasp, and final frames from successful \method{} rollouts. Single-arm rows (\emph{stack cups}, \emph{pick and place}, \emph{remove cup cap}) are side-view recordings cropped to the workspace; bimanual rows (\emph{stack cups}, \emph{bowl}) are frames from the policy's top camera.}
\label{fig:real_rollouts}
\end{figure*}

\paragraph{Rollouts and failure analysis.}
Figure~\ref{fig:real_rollouts} shows keyframes from successful \method{}
rollouts. On tasks with large object moves, the lookahead gives a clear target
for the reach: single-arm \emph{stack cups} and \emph{remove cup cap} reach
$94\%$ and $88\%$, and the bimanual tasks $84\%$ and $72\%$ (vs.\ $66\%$ and
$64\%$ for $\pi_{0.5}$). On \emph{pick and place} ($58\%$), most failures occur
at the final grasp. With the VAE's $8\times$ spatial stride and $2\times$
patchification, the cube covers only 1--2 latent tokens, so the lookahead
guides the arm toward the cube but cannot resolve sub-centimeter finger
alignment; $\pi_{0.5}$'s robot-data pretraining gives it better local
precision here.

\clearpage

\section{Reproducibility and training details}
\label{app:repro}

\paragraph{Backbone and latent space.}
\method{} builds on SkyReels-V2-DF (1.3B)~\citep{skyreels2025}.
A raw video sequence $\mathbf{V} \in \mathbb{R}^{T \times H \times W \times 3}$ is compressed into a continuous latent representation $\mathbf{Z} \in \mathbb{R}^{T' \times H' \times W' \times C}$ by the spatiotemporal causal video VAE~\citep{wan2025,skyreels2025} with temporal downsampling ratio $p_t=4$, spatial downsampling ratio $p_s=8$, and latent dimension $C=16$; the latents are then patchified with patch size $p_h = p_w = 2$, yielding $N = (H/16) \times (W/16)$ spatial tokens per frame.
Language conditioning enters the DiT via cross-attention, while per-frame noise timesteps $\tau$ modulate intermediate features via adaptive layer normalization (adaLN)~\citep{dit2023}.

\paragraph{Hyperparameters and training configuration.}
All models are trained with PyTorch on $4\times$ NVIDIA H200 (141\,GB SXM5) GPUs using \texttt{bfloat16} mixed precision. We optimize the model using AdamW ($\beta_1=0.9, \beta_2=0.95, \epsilon=10^{-8}$, weight decay $10^{-8}$) with gradient norm clipping at $1.0$. The video DiT backbone is trained with a base learning rate of $1.0\times 10^{-5}$, while the action head is trained with a learning rate of $1.0\times 10^{-4}$, both scheduled via cosine decay with 5000 warmup steps down to a minimum learning rate of $5.0\times 10^{-7}$. Per-device batch size is 16 ($64$ total batch size). The action head cross-attends to DiT layers $\{5, 12, 19, 26\}$ with dropout probability $p=0.2$. The lookahead dropout probability is $p=0.1$. Flow timesteps are not sampled uniformly: video targets use $\tau = 5u/(1+4u)$ with $u = \mathrm{sigmoid}(n)$, $n \sim \mathcal{N}(0, 1)$ (logit-normal with timestep shift $5$)~\citep{sd3_2024}, and actions use $\sigma = 1 - s + s\,x$ with $x \sim \mathrm{Beta}(1.5, 1)$ and $s = 0.999$~\citep{pi0_2024}. On RoboCasa, the action loss is additionally reweighted by a bell-shaped function of $\sigma$ that peaks at $0.5$, vanishes at both ends, and has unit mean; LIBERO uses the unweighted loss. Checkpoints are evaluated using exponential moving average (EMA) with decay rate $0.999$.

\end{document}